\def\year{2019}\relax
\documentclass[letterpaper]{article} %DO NOT CHANGE THIS
\usepackage{aaai19}  %Required
\usepackage{times}  %Required
\usepackage{helvet}  %Required
\usepackage{courier}  %Required
\usepackage{url}  %Required
\usepackage{graphicx}  %Required

\usepackage{tikz-dependency}
\usepackage{lipsum,adjustbox}
\usepackage{amsfonts}
\usepackage{amsmath}
\usepackage{multirow}
\usepackage{subcaption}
\usepackage{url}
\usepackage{epsfig}
\usepackage{fixltx2e}
\usepackage{array}
\usepackage{booktabs}

\begin{document}
% The file aaai.sty is the style file for AAAI Press 
% proceedings, working notes, and technical reports.
%
\title{One for All: Neural Joint Modeling of Entities and Events}

%\author{AAAI Press\\
%Association for the Advancement of Artificial Intelligence\\
%2275 East Bayshore Road, Suite 160\\
%Palo Alto, California 94303\\
%}

%\author{
%Trung Minh Nguyen$^\dag$ and Thien Huu Nguyen$^\#$$^\ddagger$ \\
% $^\dag$ Alt Inc., 8F, Higashi-Kanda 3-1-2, Chiyoda-ku, Tokyo, 101-0031, Japan \\
% $^\#$ Montreal Institute for Learning Algorithms, University of Montreal, Canada \\
% $^\ddagger$ Department of Computer and Information Science, University of Oregon, USA \\
% {\small \tt 
%nguyen.minh.trung@alt.ai,thien@cs.uoregon.edu} }

\author{ 
  Trung Minh Nguyen\thanks{Equal contribution.} \\
  Alt Inc. \\
  8F, Higashi-Kanda 3-1-2, Chiyoda-ku \\
  Tokyo, 101-0031, Japan \\
  {\tt nguyen.minh.trung@alt.ai} \\
  \And
  Thien Huu Nguyen\footnotemark[1] \\
  Department of Computer and Information Science \\
  University of Oregon \\
  Eugene, Oregon 97403, USA \\
  {\tt thien@cs.uoregon.edu}}

\maketitle
\begin{abstract}
The previous work for event extraction has mainly focused on the predictions for event triggers and argument roles, treating entity mentions as being provided by human annotators. This is unrealistic as entity mentions are usually predicted by some existing toolkits whose errors might be propagated to the event trigger and argument role recognition. Few of the recent work has addressed this problem by jointly predicting entity mentions, event triggers and arguments. However, such work is limited to using discrete engineering features to represent contextual information for the individual tasks and their interactions. In this work, we propose a novel model to jointly perform predictions for entity mentions, event triggers and arguments based on the shared hidden representations from deep learning. The experiments demonstrate the benefits of the proposed method, leading to the state-of-the-art performance for event extraction.
\end{abstract}

\section{Introduction}

An important problem of information extraction in natural language processing (NLP) is event extraction (EE): understanding how events are presented in text and developing techniques to recognize such events. We follow the definition of events in the annotation guideline designed for the ACE 2005 dataset\footnote{\tiny \url{https://www.ldc.upenn.edu/collaborations/past-projects/ace}}: an event is triggered by some words in the sentence with which several entities are associated to play different roles in the event.

%an event is something that happens or leads to some change of state. Concretely, it

EE is a challenging problem as it is a composition of three subtasks corresponding to different aspects of the event definition. In particular, the first subtask concerns the extraction of entity mentions appearing in the sentences (Entity Mention Detection - EMD) while the second subtask needs to identify the event trigger words (Event Detection - ED). Finally, in the third subtask, the relationships between the detected entity mentions and trigger words in the sentences should be recognized to reflect the roles of the entity mentions in the events (Argument Role Prediction - ARP). We call the three subtasks ordered by ``EMD $\rightarrow$ ED $\rightarrow$ ARP'' the EE pipeline for convenience. For instance, consider the following sentence taken from the ACE 2005 dataset:

{\it Another a-10 warthog was hit today.}

In this sentence, an EMD system needs to recognize ``{\it a-10 warthog}'' as an entity mention of type ``{\it VEHICLE}'' and ``{\it today}'' as a time expression. For ED, the systems should be able to realize that ``{\it hit}'' is a trigger word for an event of type ``{\it Attack}''. Finally, for ARP, the systems are supposed to identify ``{\it a-10 warthog}'' as playing the ``{\it Target}'' role in the ``{\it Attack}'' event and ``{\it today}'' as the event's time. 

%Each of these three subtasks is already challenging on its own due to the flexibility and ambiguity of natural languages \cite{Nguyen:16a}. It becomes even more challenging as these subtasks are assembled into a single task, for which some dependencies/interactions between the subtasks arises and should be modeled \cite{Li:13}. We call the three subtasks ordered by ``EMD $\rightarrow$ ED $\rightarrow$ ARP'' the EE pipeline for convenience.

%For instance, many of the prior work are dedicated to solving only EMD or ED \cite{Chen:15}. Some recent studies have attempted to jointly model ED and ARP \cite{Li:13,Nguyen:16a} that requires manual annotation for entity mentions to be provided.

%Presumably, if these separate systems were implemented in practice, the outputs from an earlier system in the EE pipeline would be fed into the later systems as a possible input.

%For instance, it has been shown in \cite{Li:13} that feeding the predicted entity mentions of a separate EMD system into a joint system for ED and ARP would lead to significant performance drops for trigger and argument recognition, comparing to the setting where perfect entity mentions are employed instead.

A large portion of the prior work on EE has taken a simplified approach that only focuses on one or two specific subtasks, either assuming the manual/golden annotation for the other subtasks or simply ignoring them (i.e, the pipelined approach) \cite{Li:13,Chen:15,Nguyen:16a}. One of the major issues with this approach is the error propagation in which the error from the earlier subtask is inherited and magnified in the later subtasks, causing the poor performance of those later subtasks \cite{Li:13}. In addition, the pipelined approach for EE does not have any mechanisms to capture the dependencies and interactions among the three subtasks so the later subtasks in the pipeline can interfere with and improve the decision process for the earlier subtasks. The earlier subtasks, on the other hand, can only communicate with the later subtasks via the discrete outputs, and fail to pass deeper information to the later stages to potentially improve the overall performance. Consider an EE system where EMD is done separately from ED and ARP as an example. In this system, the EMD module would work on its own and the ED and ARP modules have no way to correct the mistake made earlier by the EMD module. At the same time, it is very usual that the EMD module can only provide the ED and ARP modules with the boundary and types of the detected entity mentions. Such deeper information as the hidden contextual representations or the more fine-grained semantic classification of the entity mentions cannot be passed to or affect the ED and ARP modules. This would cause the inefficiency to use the information among the subtasks and result in the poor performance for EE.

%involve the information networks for entities, relations and events in \cite{Li:14}, and the joint framework to extract entities and events in \cite{Yang:16}

%To the best of our knowledge, t

It is thus appealing to design a single system to simultaneously model the three EE subtasks to avoid the aforementioned issues of the pipelined approach. However, due to the complexity in modeling, there has been only few works in the literature to study this joint modeling approach for EE. The major prior works in this direction for the ACE 2005 dataset involve \cite{Li:14}, \cite{Alex:16} and \cite{Yang:16}. Although these studies address the issues associated with the separate approach to some extent, they share the same limitation in which binary features (i.e, lexical words, dependency paths, etc.) are the main tools to capture the context for the individual subtasks and the dependencies/interactions among them. The major issue of those binary features is the inability to generalize over unseen words/features (due to the hard matches of binary features) and the limited expressiveness to encode the effective hidden structures for EE \cite{Nguyen:16a}. Specifically, such binary representations cannot take advantages of the deep learning (DL) models with the shared hidden representations across different stages, a useful mechanism to enable the communications among the subtasks for EE demonstrated in \cite{Nguyen:16a}.

In order to overcome the issues of such prior works for EE, in this paper, we propose a single deep learning model to jointly solve the three subtasks EMD, ED and ARP of EE. In particular, we employ a bidirectional recurrent neural network (RNN) to induce the shared hidden representations for the words in the sentence, over which the predictions for all the three subtasks EMD, ED and ARP are made. On the one hand, the bidirectional RNN helps to induce effective underlying structures via real-valued representations for the EE subtasks and mitigate the issue of hard matches for binary features. On the other hand, the shared hidden representations for the three subtasks enable the knowledge sharing across the subtasks so the hidden dependencies/interactions of the subtasks can be exploited to improve the EE performance.

%improving the generalization over unseen words/features

%Finally, we also investigate the mechanisms to inject the predictions of the earlier stages as additional features for the later stages in the the joint model.

We conduct extensive experiments to evaluate the effectiveness of the proposed model. The experiments demonstrate the benefits of joint modeling with deep learning for the three subtasks of EE over the traditional baselines, yielding the state-of-the-art performance on the long-standing and widely-used dataset ACE 2005. To the best of our knowledge, this is the first work to jointly model EMD, ED and ARP with deep learning.

%\footnote{Our code will be available after publication.}

%In addition to this communication channel with the hidden representations at the context input level, we also explicitly circulate the predictions of the earlier subtasks

\section{Related Work}

The early work on EE has mainly focused on the pipelined approach that performs the subtasks for EE separately and heavily relies on feature engineering to extract a diversity of features \cite{Grishman:05,Ahn:06,Ji:08,Gupta:09,Patwardhan:09,Liao:10,Liao:11,Hong:11,McClosky:11,Huang:12,Makoto:14,Li:15}. Some recent work has developed joint inference models for ED and ARP to address the error propagation issue in the pipelined approach. Those work exploits different structured prediction methods, including Markov Logic Networks \cite{Riedel:09,Poon:10,Venugopal:14}, Structured Perceptron \cite{Li:13,Li:14,Alex:16} and Dual Decomposition \cite{Riedel:09,Riedel:11a,Riedel:11b}. The closest work to ours is \cite{Yang:16} that attempts to jointly model EMD, ED and ARP for EE. However, this work needs to find entity mentions and event trigger candidate separately. It also does not employ shared hidden feature representations as we do in this work with DL.

Deep learning has been shown to be very successful for EE recently. Most of the early work in this direction has also followed the pipelined approach \cite{Nguyen:15b,Chen:15,Nguyen:16d,Chen:17,Liu:17,Nguyen:18a,Liu:18,Huang:18,Lu:18} while some work on joint inference for EE has also been introduced \cite{Nguyen:16a,Sha:18}. However, these studies are limited to the joint modeling of ED and ARP only.

\section{Model}

We propose a joint model for the three subtasks of EE (i.e, EMD, ED and ARP) at the sentence level. Let $W=w_1,w_2,\ldots,w_n$ be a sentence with $n$ as the number of words/tokens and $w_i$ as the $i$-th token. In order to solve the EMD problem, we cast it as a sequence labeling problem that attempts to assign a label $e_i$ for every word $w_i$ in $W$. The result is a label sequence $E = e_1,e_2,\ldots,e_n$ for $W$ that can be used to reveal the boundary of the entity mentions and their entity types in the sentence. We apply the BIO annotation schema to generate the BIO labels for the words in the sentences\footnote{Following \cite{Yang:16}, we consider values and time expressions as two additional entity types for prediction in the ACE 2005 dataset.}. %, leading to a set of 9 entity types in this work, i.e, PER, ORG, GPE, LOC, FAC, VEH, WEA, VALUE and TIME.}.

\begin{figure*}[!htb]
\centering
\addtolength{\abovecaptionskip}{-3.0mm}
\addtolength{\belowcaptionskip}{-5mm}
\includegraphics[scale=0.485]{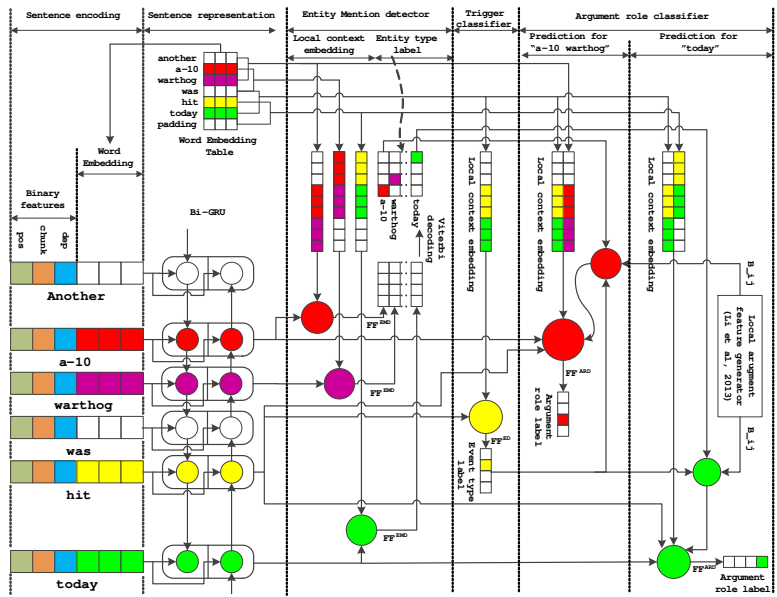}
\caption{\small The joint EE model for the three subtasks with the input sentence ``{\it Another a-10 warthog was hit today}'' with local context window $u=1$. Red and violet correspond to the beginning and last tokens of the entity mention ``{\it a-10 warthog}'' while green corresponds to the time ``{\it today}''. The trigger candidate ``{\it hit}'' at the current token is associated with yellow.}
\label{fig:model}
\end{figure*}

%Note that as we have 9 entity types in total, the BIO annotation schema would render a 19-class classification for EDM for the words in $W$ (i.e, one class is reserved for the tokens that do not belong to any entity mentions).

Regarding the ED task for triggers, we follow the prior works \cite{Li:13,Nguyen:16a,Sha:18} to assume event triggers to be only single words/tokens in the sentences. This essentially leads to a word classification problem for every word in the sentences in which we need to predict an event type $t_i$ for $w_i \in W$ ($t_i$ can be ``{\it Other}'' to indicate the word $w_i$ is not triggering any event of interest). The sequence of event type labels for the words in $W$ is denoted by $T=t_1,t_2,\ldots,t_n$.

Finally, for event arguments, we need to recognize the entity mentions that are arguments for the event mentions appearing in $W$. However, as the event mentions and triggers are not provided in advanced in our setting, we essentially need to predict the argument role label for every pair of entity mention candidates and trigger candidates in the sentence. We choose the indexes of the beginning tokens of the entity mentions as the single anchors for the entity mentions.  This translates into an argument role label matrix $A=\left(a_{ij}\right)_{i,j=1}^n$ to encode the argument information of the events in $W$. In this square matrix, $a_{ij}$ is set to ``{\it Other}'' if any of the following conditions is satisfied: (i) $i = j$, (ii) $w_i$ is not a trigger word for any events in $W$, and (iii) $w_j$ is not the beginning token of any entity mentions in $W$. Otherwise, if all the conditions are not satisfied, $a_{ij}$ will be the argument role label that the entity mention with the beginning token $w_j$ has in the event mention associated with the trigger word $w_i$. For convenience, we denote $a_i$ as the $i$-th row in the matrix $A$. Given this encoding schema, the goal of the ARP module in our model is to predict the labels for the elements $a_{ij}$ of $A$ using the context specific to the tokens $w_i$ and $w_j$.

%Note that even in the second case, $a_{ij}$ can still be ``{\it Other}'' if the entity mention with $w_j$ does not have any role in the event mention trigged by $w_i$.

%For clarification, the notations $E=e_1,e_2,\ldots,e_n$, $T=t_1,t_2,\ldots,t_n$, and $A=\left(a_{ij}\right)_{i,j=1}^n$ above are the random variables to indicate the labels. For the actual predicted labels, we will the 

%The overall architecture of the joint model for EE involves two components (i.e, encoding and decoding). In the encoding component, the input sentence $W$ is transformed into a hidden representation while in the decoding component, the hidden representation is used to make predictions for the three subtasks EMD, ED and ARP for EE. Figure \ref{fig:model} shows an overview of the proposed model.

The overall architecture of the joint model for EE in this work involves five components : i.e, sentence encoding, sentence representation, entity mention detector, trigger classifier and argument role classifier. These components are chained in an order as demonstrated in Figure \ref{fig:model} (from left to right). The first two steps help to transform the input sentence $W$ into a hidden representation while the last three steps consume this hidden representation to make predictions for the three subtasks EMD, ED and ARP of EE.

%In our model, the hidden representation is shared across all the three subtasks to facilitate the communication among the subtasks in the joint model.

%In the following, we will first describe an end-to-end version of the two components for the joint EE model that only take the sentence $W$ and the word embeddings as the inputs. We would then describe the additional features that are helpful for our model.

\subsection{Sentence Encoding}
\label{sec:enc}

In the first component of {\it sentence encoding}, every word $w_i \in W$ is transformed into a vector $x_i$ using the concatenation of the following vectors:

1. The pre-trained word embedding $d_i$ of $x_i$ \cite{Mikolov:13a}. We update the pre-trained word embeddings during the training process.

%These features are motivated by the previous work on EE and named entity recognition .

2. The binary vectors to capture the POS, chunk, and dependency information for $w_i$ in $W$ \cite{Nguyen:16a}. In particular, we first run a POS tagger, chunker and dependency parser over the input sentence $W$. The results are then used to collect the POS tag, chunking tag (with the BIO annotation schema) and the surrounding dependency relations in the parse tree for $w_i$. Finally, we create one-hot vectors to represent the POS tag and chunking tag for $w_e$ as well as a binary vector to indicate which dependency relations surround $w_i$ in the dependency tree.

\subsection{Sentence Representation}

After the sentence encoding step, the input sentence $W$ becomes a sequence of vectors $X=x_1,x_2,\ldots,x_n$. In the {\it sentence representation} component, the sequence of vectors $X$ is fed into a bidirectional recurrent neural network \cite{Hochreiter:97,Cho:14} to generate the hidden vector sequence $H=h_1,h_2,\ldots,h_n$ for every word in the $W$. We employ the Gated Recurrent Units (GRU) \cite{Cho:14} to implement the RNN model in this work. It has been shown that the hidden vector sequence $h_1,h_2,\ldots,h_n$ encodes rich contextual information of the whole sentence in each of the hidden vector $h_i$ for EE \cite{Nguyen:16a}. It is important to note that we utilize $H$ as the shared representation to make the predictions for all the following components for EMD, ED and ARP.  This enables the communications and facilitates the knowledge transfer among the three subtasks.

%Note that each hidden vector $h_i$ is the concatenation of the hidden vectors obtained from the forward and backward RNNs over $X$. 

In order to decode $W$ for EE, our goal is to predict the label variables in $E$, $T$ and $A$ jointly. Formally, this amounts to estimating the joint probability $P(A,T,E|W)$ for the input sentence $W$. In this work, we decompose this probability as follows to guide our design of the model architecture:
\begin{equation*}
\small
\label{eq:prob}
\begin{split}
 P(A,T,E|W) & =  P(E|W) \times P(A,T|E,W) \\
			&  = P(E|W) \times P(a_1,t_1|E,W) \\
			& \times P(a_2,t_2|E,W,a_{<2},t_{<2}) \\
			& \ldots \\
			& \times P(a_n,t_n|E,W,a_{<n}, t_{<n}) \\
\end{split}
\end{equation*}
where $a_i$ denote the $i$-th row in the argument role label matrix $A$ and $t_{<i} = t_1,t_2,\ldots,t_{i-1}$, $a_{<i} = a_1,a_2,\ldots,a_{i-1}$.

Based on this decomposition, we would first predict the entity type label $e_i$ for every word in the sentence (i.e, computing $P(E|W)$ for EMD) in the {\it entity mention detector} component. Afterward, the sentence is scanned from left to right for which the probability $P(a_i,t_i|E,W,a_{<i},t_{<i})$ is estimated at the step/word $i$ for trigger and argument predictions (i.e, in the last two components {\it trigger classifier} and {\it argument role classifier} of the overall architecture). We describe how those probabilities are computed using the hidden vectors $h_i$ and the word embeddings $d_i$ in the following. Note that the modeling of $P(a_i,t_i|E,W,a_{<i},t_{<i})$ enables the use of the information from $a_{<i}$ and $t_{<i}$ to reveal the inter-dependencies among the multiple events appearing in the input sentence to better predict $a_i$ and $t_i$.

Throughout this paper, we will refer to ``{\it the local context $D_i$ of a token $w_i$} in $W$'' as the concatenation vector of the word embeddings of the words in a window $u$ of $w_i$ in $W$, i.e: $D_i = [d_{i-u},\ldots,d_i,\ldots,d_{i+u}]$ where zero vectors are padded if the index is out of range.

\subsubsection{Entity Mention Detector}

For entity mention prediction, the probability $P(E|W)$ can be decomposed into:
\begin{equation*}
\small
\begin{split}
P(E|W)  = P(e_1|W)P(e_2|W,e_{<2}) \ldots P(e_n|W,e_{<n})
\end{split}
\end{equation*}
where $e_{<t}=e_1,e_2,\ldots,e_{i-1}$.

In this work, for each word $w_i$, we estimate $P(e_i|W,e_{<i})) = FF^{EMD}(R_i^{EMD})$ where $FF^{EMD}$ is a feed-forward neural network followed by a softmax layer to transform the feature representation $R_i^{EMD}$ for $w_i$ in EMD into a probability distribution over the possible entity type labels for $w_i$. The feature representation $R_i^{EMD}$ is, in turn, computed by concatenating the hidden vector $h_i$ and the local context $D_i$ for $w_i$: $R_i^{EMD} = [h_i, D_i]$.

%\begin{equation*}
%R_i^{EMD} = [h_i, D_i]
%\end{equation*}
Note that in the $R_i^{EMD}$ representation for $w_i$, we do not use any information about the entity type prediction made for $w_{i-1}$ as we find it not effective for our joint model from the development experiments. However, this might cause the orphan label issue (i.e, the I (inside) label for an entity type is not preceded by the B (beginning) label of the corresponding entity type). We prevent this issue by generating a transition score matrix between the entity type labels that penalizes any transitions to an I label but not from the corresponding B label. The Viterbi decoding algorithm is then employed to find the best predicted entity label sequence $E^P = e^p_1,e^p_2,\ldots,e^p_n$ for $W$ based on the scores $P(e_i|W,e_{<i}))$ ($1 \le i \le n$) and the generated transition matrix \cite{Ma:16,He:17}.

\subsubsection{Trigger and Argument Prediction}

Once the entity type label for every word in $W$ has been decided, we continue with the predictions of event triggers and arguments in the {\it trigger classifier} and {\it argument role classifier} components. As mentioned above, this step is done sequentially over the sentence from left to right. At the current word/step $i$, we attempt to compute the probability $P(a_i,t_i|E,W,a_{<i},t_{<i})$ that is decomposed into:
\begin{equation*}
\small
\begin{split}
P(a_i,t_i|E,W,a_{<i},t_{<i}) & = P(t_i|E,W,a_{<i},t_{<i}) \\ 
& \times P(a_{i1}|E,W,a_{<i},t_{<i+1}) \\
& \times P(a_{i2}|E,W,a_{i,<2},a_{<i},t_{<i+1}) \\
& \ldots \\
& \times P(a_{in}|E,W,a_{i,<n},a_{<i},t_{<i+1})
\end{split}
\end{equation*}
where $a_{i,<j}=a_{i,1},a_{i,2},\ldots,a_{i,j-1}$.

In this product, the term $P(t_i|E,W,a_{<i},t_{<i})$ is to predict the event type that the current word $w_i$ is triggering. Note that this can output the ``{\it Other}'' type to indicate that the current word is not an event trigger. The term $P(a_{ij}|E,W,a_{i,<j},a_{<i},t_{<i+1})$, on the other hand, predict the role that the entity mention with the beginning token of $w_j$ plays in the event mention associated with $w_i$ (i.e, the current event mention). Note that $P(a_{ij}|E,W,a_{i,<j},a_{<i},t_{<i+1})$ is only meaningful if $w_i$ is a trigger word and $w_j$ is the beginning token of some entity mention in the sentence. In other cases, we can simply skip the computation for $P(a_{ij}|E,W,a_{i,<j},a_{<i},t_{<i+1})$. During the training phase, we use the golden entity mentions to decide which tokens start an entity mention while in the evaluation phase, the predicted entity type labels $E^P = e^p_1,e^p_2,\ldots,e^p_n$ in the previous step is used for this purpose.

In order to produce $P(t_i|E,W,a_{<i},t_{<i})$, we compute the feature representation $R^{ED}_i$ for the current word $w_i$ feed it into a feed-forward neural network with softmax $FF^{ED}$, resulting in a probability distributions over the possible event types: $P(t_i|E,W,a_{<i},t_{<i}) = FF^{ED}(R^{ED}_i)$. Similar to EMD, we also compute the representation $R^{ED}_i$ by: $R_i^{ED} = [h_i, D_i]$.

%\begin{equation*}
%R_i^{ED} = [h_i, D_i]
%\end{equation*}
A greedy decoder is applied to decide the predicted event type $t^p_i$ for the current word: $t^p_i = \mbox{argmax }P(t_i|E,W,a_{<i},t_{<i})$.

Regarding the argument role distribution $P(a_{ij}|E,W,a_{i,<j},a_{<i},t_{<i+1})$, we also compute it with the feed-forward network $FF^{ARP}$ and the feature representation $R^{ARP}_{ij}$: $P(a_{ij}|E,W,a_{i,<j},a_{<i},t_{<i+1}) = FF^{ARP}(R^{ARP}_{ij})$. However, $R^{ARP}_{i,j}$ in this case is computed as:
\begin{equation}\label{rep:arg}
\small
R_{ij}^{ARP} = [h_i, D_i, h_j, D_j, V(e^p_i), V(t^p_j), M_i, B_{ij}]
\end{equation}
where $V(x)$ is the function that converts a label $x$ into an one-hot vector to represent the label. Note that during the training process, $e^p_i$ and $t^p_i$ would be set to the golden labels from the training data. $M_i$ is the binary vector to indicate the event types and argument roles that appear before step $i$ in $W$. Finally, $B_{ij}$ is the binary vector inherited from \cite{Li:13,Nguyen:16a} to capture the discrete structures/features for argument prediction between the tokens $i$ and $j$ in the sentence (i.e, the shortest dependency paths, the context words etc.). Note that different from the prior work \cite{Li:13,Nguyen:16a}, $B_{ij}$ does not contain any features related to the entity types or subtypes of the entity mentions as these features are not available to us at the beginning. We instead resort to the predicted entity type $e^p_i$ as demonstrated in Equation \ref{rep:arg}. We also apply the greedy strategy to predict the argument role $a^p_{ij}$ in this case: $a^p_{ij} = \mbox{argmax } P(a_{ij}|E,W,a_{i,<j},a_{<i},t_{<i+1})$. This completes the description of our joint model for EE.

\subsection{Training}

%In order to train the joint model, the usual way is to optimize the negative log-likelihood function:
%\begin{equation*}
%\small 
%\begin{split}
%& C(A,T,E,W) = - \log P(A,T,E|W) =  \\
%& -\sum_{i=1}^n \log P(e_i | W, e_{<i}) \\
%& - \sum_{i=1}^n \log P(t_i | E, W, a_{<i}, t_{<i}) \\
%& - \sum_{i=1}^n \sum_{j=1}^n \log P(a_{ij}|E,W,a_{i,<j},a_{<i},t_{<i+1}) \\
%& - \log P(E|W) - \log P(T|W,E) - \log P(A|W,E,T)
%\end{split}
%\end{equation*}
%However, in practice, we find that it is helpful to penalize the terms for the subtasks EMD, ED and ARP appropriately to encourage them to converge at the same time. Consequently, in this work, we instead optimize the following function to train the model:

We train the joint model by optimizing the negative log-likelihood function $C(A,T,E,W) = - \log P(A,T,E|W)$. In order to encourage the loss terms for EMD, ED and ARP to converge at the same time, we penalize these terms differently in $C(A,T,E,W)$, leading to the following loss function in this work:

\begin{equation*}
\small 
\begin{split}
& C^*(A,T,E,W) =   \\
& -\alpha\sum_{i=1}^n \log P(e_i | W, e_{<i}) \\
& - \beta\sum_{i=1}^n \log P(t_i | E, W, a_{<i}, t_{<i}) \\
& - \gamma\sum_{i=1}^n \sum_{j=1}^n \log P(a_{ij}|E,W,a_{i,<j},a_{<i},t_{<i+1}) \\
%& - \log P(E|W) - \log P(T|W,E) - \log P(A|W,E,T)
\end{split}
\end{equation*}

%\begin{equation*}
%\small
%\begin{split}
%& C*(A,T,E,W) =  - \alpha \log P(E|W) \\
%&  - \beta \log P(T|W,E) - \gamma \log P(A|W,E,T)
%\end{split}
%\end{equation*}

where $\alpha$, $\beta$ and $\gamma$ are the hyper-parameters. We use the SGD algorithm to optimize the parameters with mini-batches and the \textit{Adadelta} update rules \cite{Zeiler:12}. The gradients are computed with back-propagation while the parameters are rescaled if their Frobenius norms exceed a hyper-parameter.

\section{Experiments}

\subsection{Dataset, Parameters, and Resources}

%(672 sentences), (836 sentences), (14,849 sentences)

We evaluate the proposed model on the ACE 2005 dataset. In order to ensure a fair comparison, we use the same data split with the prior work on this dataset \cite{Li:13,Nguyen:16a,Nguyen:16b,Yang:16,Sha:18} in which 40 newswire documents are used for the test set, 30 other documents are reserved for the development set, and the remaining 529 documents form the training set. We utilize the Stanford CoreNLP to do the pre-processing for the sentences (i.e, POS tagging, chunking and dependency parsing). The pre-trained word embeddings are obtained from \cite{Nguyen:16a}.

%that pre-train the embeddings using the concatenation variant of the CBOW method in \cite{Mikolov:13a,Mikolov:13b}.

\begin{table*}[htbp]
\addtolength{\abovecaptionskip}{-3.0mm}
\addtolength{\belowcaptionskip}{-5.0mm}
\centering
\small
\resizebox{\textwidth}{!}{
\begin{tabular}{l|ccc|ccc|ccc|ccc}
\toprule
Model & \multicolumn{3}{c|}{Event Trigger} & \multicolumn{3}{c|}{Event Trigger} & \multicolumn{3}{c|}{Event Argument} & \multicolumn{3}{c}{Argument Role}  \\
 & \multicolumn{3}{c|}{Identification} & \multicolumn{3}{c|}{Classification} & \multicolumn{3}{c|}{Identification} & \multicolumn{3}{c}{Classification}  \\ 
%\cline{2-13}
\cmidrule{2-13}
 & P & R & F & P & R & F & P & R & F & P & R & F \\ 
\midrule
{\it StagedMaxent} & 73.9 & 66.5 & 70.0 & 70.4 & 63.3 & 66.7 & 75.7 & 20.2 & 31.9 & 71.2 & 19.0 & 30.0 \\
{\it Pipelined-Feature} & 76.6 & 58.7 & 66.5 & 74.0 & 56.7 & 64.2 & 74.6 & 25.5 & 38.0 & 68.8 & 23.5 & 35.0 \\
{\it Pipelined-Deep-Learning} & 72.7 & 65.9 & 69.1 & 70.4 & 63.9 & 67.0 & 61.7 & 42.1 & 50.1 & 46.0 & 31.4 & 37.4 \\
{\it Joint-Feature-Sentence} & 76.9 & 63.8 & 69.7 & 74.7 & 62.0 & 67.7 & 72.4 & 37.2 & 49.2 & 69.9 & 35.9 & 47.4 \\
{\it Joint-Feature-Document\dag} & 77.6 & 65.4 & 71.0 & 75.1 & 63.3 & 68.7 & 73.7 & 38.5 & 50.6 & 70.6 & 36.9 & 48.4 \\
{\it NP-Candidate-Deep-Learning} & - & - & - & - & - & 69.6 & - & - & 57.2 & - & - & 50.1 \\ \hline
{\it Joint3EE} & 70.5 & 74.5 & {\bf 72.5} & 68.0 & 71.8 & {\bf 69.8} & 59.9 & 59.8 & {\bf 59.9} & 52.1 & 52.1 & {\bf 52.1} \\
\bottomrule
\end{tabular}
}
%\caption{Performance on the ACE 2005 test set\protect\footnotemark.}
\caption{Performance on the ACE 2005 test set. The comparison between {\it Joint3EE} and {\it Pipelined-Deep-Learning} is significant with $p < 0.05$. ``\dag'' designates the systems with document level information.}
\label{tab:main}
\end{table*}

Regarding the hyper-parameters, the word embeddings have the dimension of 300; the number of hidden units in the encoding RNNs is 300; and the window for local context $u$ is 2. We use the feed-forward neural networks with one hidden layer of 600 hidden units for $FF^{EMD}$, $FF^{ED}$ and $FF^{ARP}$. The mini-batch size is 50 while the Frobenius  norm for the parameters norms is 3. These values give us best the results on the development set. For the penalty coefficients in the objective function, the best values we obtained from the development data is $\alpha=0.5, \beta=1.0, \gamma=0.5$. We also implement dropouts on the input word embeddings and the hidden vectors of the feed-forward networks with a rate of 0.5 (tuned on the development set). Finally, the same correctness criteria with the previous work \cite{Nguyen:16a,Yang:16,Sha:18} is applied when we evaluate the predicted results.

\subsection{Comparing to the State of the Art for Trigger and Argument Predictions}

% \begin{table*}[htbp]
% %\addtolength{\abovecaptionskip}{-2.0mm}
% \addtolength{\belowcaptionskip}{-5.0mm}
% \centering
% \resizebox{\textwidth}{!}{
% \begin{tabular}{|l|c|c|c|c|c|c|c|c|c|c|c|c|}
% \hline
% Model & \multicolumn{3}{c|}{Event Trigger} & \multicolumn{3}{c|}{Event Trigger} & \multicolumn{3}{c|}{Event Argument} & \multicolumn{3}{c|}{Argument Role}  \\
%  & \multicolumn{3}{c|}{Identification} & \multicolumn{3}{c|}{Classification} & \multicolumn{3}{c|}{Identification} & \multicolumn{3}{c|}{Classification}  \\ \cline{2-13}
%  & P & R & F & P & R & F & P & R & F & P & R & F \\ \hline
% {\it StagedMaxent} & 73.9 & 66.5 & 70.0 & 70.4 & 63.3 & 66.7 & 75.7 & 20.2 & 31.9 & 71.2 & 19.0 & 30.0 \\ \hline
% {\it Pipelined-Feature} & 76.6 & 58.7 & 66.5 & 74.0 & 56.7 & 64.2 & 74.6 & 25.5 & 38.0 & 68.8 & 23.5 & 35.0 \\ \hline
% {\it Pipelined-Deep-Learning} & 72.7 & 65.9 & 69.1 & 70.4 & 63.9 & 67.0 & 61.7 & 42.1 & 50.1 & 46.0 & 31.4 & 37.4 \\ \hline
% {\it Joint-Feature-Sentence} & 76.9 & 63.8 & 69.7 & 74.7 & 62.0 & 67.7 & 72.4 & 37.2 & 49.2 & 69.9 & 35.9 & 47.4 \\ \hline
% {\it Joint-Feature-Document} & 77.6 & 65.4 & 71.0 & 75.1 & 63.3 & 68.7 & 73.7 & 38.5 & 50.6 & 70.6 & 36.9 & 48.4 \\ \hline
% {\it Joint3EE} & 70.5 & 74.5 & {\bf 72.5} & 68.0 & 71.8 & {\bf 69.8} & 59.9 & 59.8 & {\bf 59.9} & 52.1 & 52.1 & {\bf 52.1} \\ \hline
% \end{tabular}
% }
% \caption{Performance on the ACE 2005 test set. The comparison between {\it Joint3EE} and {\it Pipelined-Deep-Learning} is significant with $p < 0.05$.}
% \label{tab:main}
% \end{table*}

In order to evaluate the effectiveness of the proposed model for event trigger and argument predictions, we compare the proposed model (called {\it Joint3EE}) with the following baselines:

1. {\it StagedMaxent}: This is a feature-based pipelined baseline presented in \cite{Yang:16} (i.e, performing the three subtasks separately). The EMD subtask is solved by a CRF tagger. All the three subtasks are based on feature engineering. % to compute feature representations.

%in the order: EMD $\rightarrow$ ED $\rightarrow$ ARP.

% that trains a structured perceptron with beam search

2. {\it Pipelined-Feature}: This is the feature-based EE system that uses the same CRF tagger as {\it StagedMaxent}'s to annotate entity mentions. The results are passed to the joint model for ED and ARP in \cite{Li:13}. Similar to the CRF tagger, this joint model also employs complicated feature engineering. This baseline is reported in \cite{Yang:16} and functions as {\it the state-of-the-art pipelined model for EE using feature engineering}.

3. {\it Pipelined-Deep-Learning}: This baseline also first extracts entity mentions and then uses the outputs in a joint model for ED and ARP (pipelined). However, the EMD model and the joint model for ED and ARP are based on deep learning in this case. In particular, the EMD model for this baseline is inherited from the EMD component of this work while the joint model for ED and ARP is provided by \cite{Nguyen:16a}. The performance of the EMD model when it is trained independently for EMD is shown in Table \ref{tab:ner} (i.e, {\it EMD-Pipelined-DL}).

%Note that the joint deep learning model in \cite{Nguyen:16a} is the state-of-the-art EE model with golden entity mentions. 

4. {\it Joint-Feature-Sentence}: This is the joint inference system that models the three subtasks of EE in a single model proposed in \cite{Yang:16}. This model only considers the structures within a single event.

5. {\it Joint-Feature-Document}: This system is similar to {\it Joint-Feature-Sentence} except that it goes beyond the sentence level and exploits the event-event dependencies at the document level \cite{Yang:16}. This is {\it the state-of-the-art joint model for EE using feature engineering}.

%It also has the best reported performance in the setting that entity mentions are predicted in our setting with the ACE 2005 dataset.

6. {\it NP-Candidate-Deep-Learning} \cite{Sha:18}: This method uses the existing tools to extract noun phrases and treat them as the argument candidates for the events. It is then followed by a dependency bridge recurrent neural network with argument interaction modeling to jointly perform ED and ARP for event extraction. This method currently has {\it the best reported performance on ED and ARP for EE among the methods that do not assume golden entity mentions with the ACE 2005 dataset}. The EMD task is not considered in this method.

%This system is similar to {\it Joint-Feature-Sentence} in that both systems perform a joint inference for EMD, ED and ARP in a single model. However, {\it Joint-Feature-Document} goes beyond the sentence level and exploits the event-event dependencies at the document level. This is {\it the state-of-the-art joint model for EE using feature engineering}. It also has the best reported performance in the setting that entity mentions are predicted for the ACE 2005 dataset.

Table \ref{tab:main} reports the performance of the systems in terms of precisions (P), recalls (R) and F1 scores (F). The first observation is that the performance of the joint deep learning model for ED and ARP in \cite{Nguyen:16a} with predicted entity mentions (i.e, {\it Pipelined-Deep-Learning} with 67.0\% and 37.4\% for trigger and argument role classification respectively) is much worse than that with perfect entity mentions in \cite{Nguyen:16a} (i.e, 69.3\% and 55.4\% for trigger and argument role classification respectively). This is consistent with the significant performance drop of the joint feature-based model for ED and ARP (reported in \cite{Li:13}) when entity mentions are predicted. Such pieces of evidence along with the significantly better performance of the fully joint models for EE (i.e, {\it Joint-Feature-Document} and {\it Joint3EE}) over the pipelined models (i.e, {\it StagedMaxEnt}, {\it Pipelined-Feature} and {\it Pipelined-Deep-Learning}) in Table \ref{tab:main} demonstrate the need to jointly perform EMD with ED and ARP to improve the EE performance. We also see that {\it Pipelined-Deep-Learning} outperforms {\it Pipelined-Feature} and {\it Joint3EE} is significantly better than {\it Joint-Feature-Document} with respect to all the F1 scores in Table \ref{tab:main}. These facts testify to the benefits of deep learning over the feature-based models for EE no matter which approach we take (i.e, pipelined or joint inference). Finally, comparing {\it Joint3EE} and the current state-of-the-art model {\it NP-Candidate-Deep-Learning}, we see that {\it Joint3EE} is superior to {\it NP-Candidate-Deep-Learning} on both trigger and argument prediction. The improvement is significant on event argument identification and argument role classification (an improvement of 2.7\% for event argument identification and 2.0\% for argument role classification on the absolute F1 scores), clearly demonstrating the effectiveness of the proposed deep learning method to jointly model the three subtasks for EE.
\subsection{Performance of EMD}

This section evaluates the EMD performance of the proposed joint model (called {\it EMD-Joint3EE}). The following baselines are chosen for comparison:

1. {\it EMD-CRF}: This is the performance of the CRF tagger for EMD used in the pipelined models {\it StagedMaxent} and {\it Pipelined-Feature}. It is implemented in \cite{Yang:16}.

2. {\it EMD-Pipelined-DL}: This is the performance of the deep learning EMD module used in {\it Pipelined-Deep-Learning} that resembles the EMD component of the proposed model {\it Joint3EE}, but is trained separately from ED and ARP.

3. {\it EMD-Joint-Feature}: This corresponds to the EMD module in {\it Joint-Feature-Document} \cite{Yang:16} that is trained jointly with ED and ARP in a single model based on feature engineering. It is currently the state-of-the-art EMD performance in the setting for EE.

\begin{table}[htbp]
\addtolength{\abovecaptionskip}{-3.0mm}
\addtolength{\belowcaptionskip}{-3.0mm}
\centering
\small
\begin{tabular}{l|ccc}
\midrule
Model & P & R & F \\
\midrule
{\it EMD-CRF} & 85.5 & 73.5 & 79.1 \\
{\it EMD-Pipelined-DL} & 80.6 & 80.3 & 80.4 \\
{\it EMD-Joint-Feature} & 82.4 & 79.2 & 80.7 \\ \hline
{\it EMD-Joint3EE} & 82.0 & 80.4 & {\bf 81.2} \\
\midrule
\end{tabular}
\caption{\small Entity Mention Detection Performance}
\label{tab:ner}
\end{table}

% \begin{table}[htbp]
% \addtolength{\abovecaptionskip}{-1.0mm}
% \addtolength{\belowcaptionskip}{-3.0mm}
% \centering
% \begin{tabular}{|l|c|c|c|}
% \hline
% Model & P & R & F \\ \hline
% {\it EMD-CRF} & 85.5 & 73.5 & 79.1 \\ \hline
% {\it EMD-Pipelined-DL} & 80.6 & 80.3 & 80.4 \\ \hline
% {\it EMD-Joint-Feature} & 82.4 & 79.2 & 80.7 \\ \hline
% {\it EMD-Joint3EE} & 82.0 & 80.4 & {\bf 81.2} \\ \hline
% \end{tabular}
% \caption{\small Entity Mention Detection Performance}
% \label{tab:ner}
% \end{table}

%\footnotetext{The comparison between {\it Joint3EE} and {\it Pipelined-Deep-Learning} is significant with $p < 0.05$.}
Table \ref{tab:ner} shows the performance of the models. First, we can see from the table that the performance of the proposed model {\it EMD-Joint3EE} is better than that of {\it EMD-Pipelined-DL}. The performance improvement is 0.8\% on the absolute F1 score and significant with $p < 0.05$. Second, we also see that {\it EMD-Joint3EE} outperforms the current state-of-the-art joint model {\it EMD-Joint-Feature} with an improvement of 0.5\% on the F1 score. Such evidence confirms the benefits of jointly modeling EMD with ED and ARP via deep learning to improve the overall performance.

 %This confirms the benefits of jointly modeling EMD with ED and ARP to improve the overall performance for EE.
 
%This helps to illustrate the advantages of doing joint inference for EMD of EE with deep learning.

%Finally, form Table \ref{tab:ner}, we see that {\it EMD-Joint3EE} outperforms the current state-of-the-art joint model {\it EMD-Joint-Feature} with an improvement of 0.5\% on the F1 score. This helps to confirm the advantages of doing joint inference for EMD of EE with deep learning.

An intriguing observation is that the performance difference for EMD between {\it Pipelined-Deep-Learning} and {\it Joint3EE} is moderate (i.e, 0.8\% in Table \ref{tab:ner}) while that difference for ARP is substantial (i.e, 13.9\% in Table \ref{tab:main}). Among several reasons, {\it Pipelined-Deep-Learning} employs the joint model for ED and ARP in \cite{Nguyen:16a} that relies on the discrete features only available to the manually annotated entity mentions such as the entity subtypes for the entity mentions (i.e, ``{\it Crime}'', ``{\it Job-Title}'' and ``{\it Numeric}'' for values). As demonstrated in Table \ref{tab:old}, such information is very helpful for ARP. However, it is not available in our setting of predicted entity mentions (i.e, only the boundaries and the entity types are predicted), causing the poor ARP performance of {\it Pipelined-Deep-Learning}.

%\begin{table}[htbp]
%\addtolength{\abovecaptionskip}{-2.0mm}
%\addtolength{\belowcaptionskip}{-3.0mm}
%\centering
%\begin{tabular}{|l|c|c|}
%\hline
%Model & Trigger & Argument \\ \hline
%\cite{Nguyen:16a} & 69.3 & 55.4 \\ \hline
%Without Entity Subtype & 64.1 & 47.5 \\ \hline
%\end{tabular}
%\caption{F1 scores for event trigger and argument classification of the joint model in \cite{Nguyen:16a} (using perfect entity mentions). The absence of the entity subtype information reduces the performance of the joint model in \cite{Nguyen:16a} by 5.2\% and 7.9\% for trigger and argument classification respectively.}
%\label{tab:old}
%\end{table}

\begin{table}[htbp]
\addtolength{\abovecaptionskip}{-1.0mm}
\addtolength{\belowcaptionskip}{-3.0mm}
\small
\centering
\begin{tabular}{l|ccc}
\toprule
Model & P & R & F \\ 
\midrule
\cite{Nguyen:16a} & 54.2 & 56.7 & 55.4 \\
Without Entity Subtype & 46.2 & 48.8 & 47.5 \\
\bottomrule
\end{tabular}
\caption{\small Performance on argument role classification of the joint model in \cite{Nguyen:16a} (using perfect entity mentions). The absence of the entity subtype information reduces the F1 score of the joint model in \cite{Nguyen:16a} by 7.9\%.}
\label{tab:old}
\end{table}

% \begin{table}[htbp]
% %\addtolength{\abovecaptionskip}{-2.0mm}
% \addtolength{\belowcaptionskip}{-3.0mm}
% \small
% \centering
% \begin{tabular}{|l|c|c|c|}
% \hline
% Model & P & R & F \\ \hline
% \cite{Nguyen:16a} & 54.2 & 56.7 & 55.4 \\ \hline
% Without Entity Subtype & 46.2 & 48.8 & 47.5 \\ \hline
% \end{tabular}
% \caption{\small Performance on argument role classification of the joint model in \cite{Nguyen:16a} (using perfect entity mentions). The absence of the entity subtype information reduces the F1 score of the joint model in \cite{Nguyen:16a} by 7.9\%.}
% \label{tab:old}
% \end{table}

%(i.e, the absence of the entity subtype information reduces the performance of the joint model in \cite{Nguyen:16a} by 5.2\% and 7.9\% for trigger and argument classification respectively)

For the {\it Joint3EE} model, although such discrete fine-grained information also does not exist explicitly, the shared hidden vectors $h_i$ across subtasks can learn to encode that information implicitly, thereby compensating the lack of information and improving the ARP performance.

%The main source of improvement comes from the precision scores, demonstrating the benefits of jointly modeling EMD with ED and ARP to correct the mistake made earlier by EMD in the EE pipeline. 

\subsection{The Effect of External Features}

There are two main sources of external features employed in {\it Joint3EE}, i.e, the binary vectors for POS, chunking and dependency parsing information in the sentence encoding, and the binary features $B_{ij}$ for argument role prediction in Equation \ref{rep:arg}. This section evaluates the effect of such features on the model performance to see how far we can reach an end-to-end model for joint EE with deep learning. Table \ref{tab:abla} presents the performance of the proposed model when such features are excluded from the model (i.e, resulting in an end-to-end model for EE (called ``{\it End-to-end-DL}'') that does not use any external hand-designed features from the NLP toolkits. We also include the performance of the state-of-the-art models (i.e, {\it Joint-Feature-Document} \cite{Yang:16} and {\it NP-Candidate-Deep-Learning} \cite{Sha:18}) that employ predicted entity mentions for EE to facilitate the comparison.
\begin{table}[htbp]
\centering

\addtolength{\abovecaptionskip}{-3.0mm}
\addtolength{\belowcaptionskip}{-3.0mm}
\resizebox{0.45\textwidth}{!}{
\begin{tabular}{l|ccc}
\toprule
Feature & Entity & Trigger & Argument \\
\midrule
{\it Joint3EE} & 81.2 & 69.8 & 52.1 \\

{\it End-to-end-DL} & 79.5 & 68.7 & 50.3 \\
\midrule
{\it Joint-Feature-Document} & 80.7 & 68.7 & 48.4 \\
{\it NP-Candidate-Deep-Learning} & - & 69.6 & 50.1 \\
\bottomrule
%- {\it Input-Fets} & 79.5 & 67.3 & 50.9 \\ \hline
%- {\it Arg-Fets} & 80.2 & 66.6 & 48.9 \\ \hline
%\begin{tabular}{@{}l@{}}- {\it Input-Fets} \\ - {\it Arg-Fets}\end{tabular} & 79.7 & 67.6 & 48.2 \\
%\bottomrule
\end{tabular}
}
\caption{\small F1 scores on classification for entity mentions (EMD), event triggers (ED) and arguments (ARP).}
\label{tab:abla}
\end{table}

% \begin{table}[htbp]
% \centering
% \addtolength{\abovecaptionskip}{-1.0mm}
% \addtolength{\belowcaptionskip}{-3.0mm}
% \begin{tabular}{|l|c|c|c|}
% \hline
% Feature & Entity & Trigger & Argument \\ \hline
% {\it Joint3EE} & 81.2 & 69.8 & 52.1 \\ \hline \hline
% - {\it Input-Fets} & 79.5 & 67.3 & 50.9 \\ \hline
% - {\it Arg-Fets} & 80.2 & 66.6 & 48.9 \\ \hline
% - {\it Input-Fets} & & & \\
% - {\it Arg-Fets} & 79.8 & 67.5 & 46.7 \\ \hline
% \end{tabular}
% \caption{\small F1 scores on classification for entity mentions (EMD), event triggers (ED) and arguments (ARP).}
% \label{tab:abla}
% \end{table}

The first observation is that the external features are useful for the joint model {\it Joint3EE} as eliminating such features downgrades the performance over all the subtasks EMD, ED and ARP (i.e. comparing {\it joint3E} and {\it End-to-end-DL}). However, the performance reduction due to this feature removal is not dramatic and the performance of the {\it End-to-end-DL} system is still comparable with that of the current state-of-the-art models {\it Joint-Feature-Document} and {\it NP-Candidate-Deep-Learning}. In particular, {\it End-to-end-DL} is only 1.2\% worse than {\it Joint-Feature-Document} on EMD and 0.9\% worse than {\it NP-Candidate-Deep-Learning} on ED. Regarding ARP, {\it End-to-end-DL} even significantly outperforms {\it Joint-Feature-Document} with 1.9\% performance improvement. These are remarkable facts given that {\it End-to-end-DL} does not use any external and manually-generated features while {\it Joint-Feature-Document} and {\it NP-Candidate-Deep-Learning} extensively rely on such external features to perform well (e.g, dependency parsing, NP chunking, gazetteers etc.). We consider this as a strong promise toward a state-of-the-art end-to-end system for EE for which the joint model in this work can be used as a good starting point.

%The main observation is that both the {\it Input-Fets} and {\it Arg-Fets} features are useful for the joint model {\it Joint3EE} as excluding any of these features worsen the performance. The {\it Input-Fets} feature seems to have strongest effects on the EMD performance as the model has the worst EMD performance when this feature is not included. Similarly, {\it Arg-Fets} seems to affect the ARP performance the most due to the substantial reduction of performance when these features are absent. When both features are taken out, we achieve an end-to-end joint model for EE that only requires word embeddings as the inputs. The performance of this end-to-end model (i.e, 79.7\% on EMD, 67.6\% on ED, and 48.2\% on ARP) is competitive to the current state-of-the-art joint model (i.e, {\it Joint-Feature-Document} with 80.7\% on EMD, 68.7\% on ED, and 48.4\% on ARP) that employed a variety of engineered external features. This demonstrates the potentials of an end-to-end joint model for EE for future research.

\subsection{Error Analysis}

\begin{table}[!htbp]
\centering
\addtolength{\abovecaptionskip}{-3.0mm}
\addtolength{\belowcaptionskip}{-3.0mm}
\begin{tabular}{lc|lc}
\toprule
\multicolumn{2}{c}{MISSED} &  \multicolumn{2}{|c}{INCORRECT} \\
Label & 			Percent 		& Label 		& Percent 	  \\ \midrule
Attack & 			16.1\% 	& End-Position 	& 18.2\% \\ 
\begin{tabular}{@{}l@{}}Transfer - \\ Ownership\end{tabular} & 12.5\%  	& Attack 		& 17.5\% \\ 
Transport & 		12.5\% 	& Transport 	& 17.5\% \\
\midrule
Total &				41.1\% 	& Total 		& 53.2\% \\
\bottomrule
\end{tabular}
\caption{\small Top three event types for trigger errors.}
\label{tab:trigger-error}
\end{table}

In order to analyze the operation of {\it Joint3EE} with respect to ED, we notice from Table \ref{tab:main} that the trigger classification performance (i.e, 69.8\%) is quite close the trigger identification performance (i.e, 72.5\%). This suggests that the main source of errors for event triggers comes from the failure to identify the trigger words. To this end, we examine the outputs of {\it Joint3EE} on the test set to determine the contributions of each event type to the trigger identification errors. Two types of errors arise in this case: (i) missing an event trigger in the test set (called {\it MISSED}), and (ii) incorrectly detecting an event trigger (called {\it INCORRECT}). Table \ref{tab:trigger-error} shows the top three event types appearing in these two types of errors and their corresponding percents over the total numbers of errors. These top three event types account for 41.1\% of the {\it MISSED} errors and 53.2\% of the {\it INCORRECT} errors. {\it Attack} and {\it Transport} are the types that are present frequently in both types of errors. A closer look at the errors reveals that the ``{\it MISSED}'' errors mostly correspond to the trigger words not appearing in the training data, such as the word ``{\it intifada}'' (of type {\it Attack}) in the following sentence:

%For example, the word ``{\it intifada}'' (of type {\it Attack}) in the following sentence of the test set does not appear in the training data:

%{\it ``It is the first time they have had freedom of movement with cars and weapons since the start of the {\bf intifada}'', the source said.}

{\it \ldots had freedom of movement with cars and weapons since the start of the {\bf intifada}'' \ldots}

The {\it INCORRECT} errors, on the other hand, belong to the confusable context that requires better modeling of the context. For instance, the word ``{\it fire}'' in the following sentence can be easily misinterpreted as an {\it Attack} event trigger by the models (due to its context with the word ``{\it car}''):

%{\it American International Group Inc. is likely to acquire all shares in GE Edison Life Insurance by the end of September and also take over GE's US car and fire insurance operations , the reports said .}

{\it \ldots also take over GE's US car and {\bf fire} insurance operations, the reports said.}

Regarding the argument prediction, we find that a large number of arguments (i.e, 209 arguments) are identified correctly, but cannot be classified properly by {\it Joint3EE} (i.e, Table \ref{tab:main}). Among those 209 arguments, there are 50 cases (23.9\%) that {\it Joint3EE} detects with the correct argument role, but assigns incorrect event types. The remaining 159 arguments (76.1\%) are associated with incorrect roles for which only 24 arguments (15.1\%) also have incorrect entity types. Consequently, the major problem for the incorrect argument classification is due to the confusion of the model on the different roles of arguments. The most frequent role confusions are between {\it Place} vs {\it Destination}, {\it Origin} vs {\it Destination}, and {\it Seller} vs {\it Buyer}. The distinction between these pairs of roles would also require better mechanisms/network architectures to model the input context.

\section{Conclusion}

We present a novel deep learning method for EE. Our model features the joint modeling of EMD, ED and ARP with the shared hidden representations across the three subtasks to enable the communications among them. We achieve the state-of-the-art performance for EE with predicted entity mentions. In the future, we plan to improve the end-to-end model so EE can be solved from just the raw sentences and the word embeddings.

\bibliographystyle{named}
\bibliography{aaai19}

\end{document}